\documentclass[hyperref]{article}

\usepackage{microtype}
\usepackage{graphicx}
\usepackage{subfigure}
\usepackage{booktabs} 

\usepackage{hyperref}

\usepackage[accepted]{icml2022}

\usepackage{amsmath}
\usepackage{amssymb}
\usepackage{mathtools}
\usepackage{amsthm}
\usepackage{amsmath,amsfonts,bm}

\usepackage[capitalize,noabbrev]{cleveref}

\theoremstyle{plain}
\newtheorem{theorem}{Theorem}[section]

\theoremstyle{definition}
\newtheorem{definition}[theorem]{Definition}

\theoremstyle{remark}

\usepackage[textsize=tiny]{todonotes}

\def\gM{{\mathcal{M}}}
\def\gR{{\mathcal{R}}}

\def\vt{{\bm{t}}}

\icmltitlerunning{On the Importance of Critical Period in Multi-stage Reinforcement Learning}

\begin{document}

\twocolumn[
\icmltitle{On the Importance of Critical Period in Multi-stage Reinforcement Learning}




\begin{icmlauthorlist}
\icmlauthor{Junseok Park}{yyy}
\icmlauthor{Inwoo Hwang}{zzz}
\icmlauthor{Min Whoo Lee}{zzz}
\icmlauthor{Hyunseok Oh}{zzz}
\icmlauthor{Minsu Lee}{comp}
\icmlauthor{Youngki Lee}{zzz}
\icmlauthor{Byoung-Tak Zhang}{yyy,zzz,comp}
\end{icmlauthorlist}

\icmlaffiliation{yyy}{Interdisciplinary Program In Neuroscience, Seoul National University, Seoul, Korea}
\icmlaffiliation{zzz}{Dept. of Computer Science and Engineering, Seoul National University, Seoul, Korea}
\icmlaffiliation{comp}{AIIS, Seoul, Korea}

\icmlcorrespondingauthor{Minsu Lee}{mslee@bi.snu.ac.kr}
\icmlcorrespondingauthor{Youngki Lee}{youngkilee@snu.ac.kr}
\icmlcorrespondingauthor{Byoung-Tak Zhang}{btzhang@bi.snu.ac.kr}

\icmlkeywords{Machine Learning, ICML}

\vskip 0.3in
]



\printAffiliationsAndNotice{}  

\begin{abstract}
The initial years of an infant's life are known as the critical period, during which the overall development of learning performance is significantly impacted due to neural plasticity. In recent studies, an AI agent, with a deep neural network mimicking mechanisms of actual neurons, exhibited a learning period similar to human's critical period. Especially during this initial period, the appropriate stimuli play a vital role in developing learning ability. However, transforming human cognitive bias into an appropriate shaping reward is quite challenging, and prior works on critical period do not focus on finding the appropriate stimulus. To take a step further, we propose multi-stage reinforcement learning to emphasize finding ``appropriate stimulus" around the critical period. Inspired by humans' early cognitive-developmental stage, we use multi-stage guidance near the critical period, and demonstrate the appropriate shaping reward (stage-2 guidance) in terms of the AI agent's performance, efficiency, and stability.

\end{abstract}

\section{Introduction}
\label{intro}

The \textit{Critical Period} is a core maturational stage of an intelligent organism's cognitive development~\cite{smart1991critical,Rice2000CriticalPO}. To learn a cognitive skill, appropriate guidance must be provided in this stage; otherwise, the organism struggles or even fails to attain the skill~\cite{nickersonCP}. In detail, neurological studies suggest that synaptic connectivity is noticeably facilitated in this stage~\cite{Takesian2013BalancingPA}. This leads to enhanced neural plasticity and sensitivity to external stimuli, and thus an organism can learn more with the same observation~\cite{nickersonCP}. For this reason, early interaction and feedback for a human baby is vital in its future development~\cite{nickersonCP,white2013learning}. 

Recent works~\cite{achille2017critical,park2021toddler,Kleijn2022critical} investigate whether the critical period only holds for biological agents and not for artificial intelligence.
These works were conducted under the intuition that, since deep neural networks are related to biological nervous systems, similar biological characteristics such as critical periods might emerge.
One study~\cite{achille2017critical} discovered the \textit{critical period phenomenon} in CNNs: neural plasticity peaks at a particular stage of training.
Other works~\cite{park2021toddler, Kleijn2022critical} found the critical period for reinforcement learning (RL) agents, and linked human guidance and interaction with critical period.

\begin{figure}[t!]
    \centering
    \includegraphics[width=0.45\textwidth]{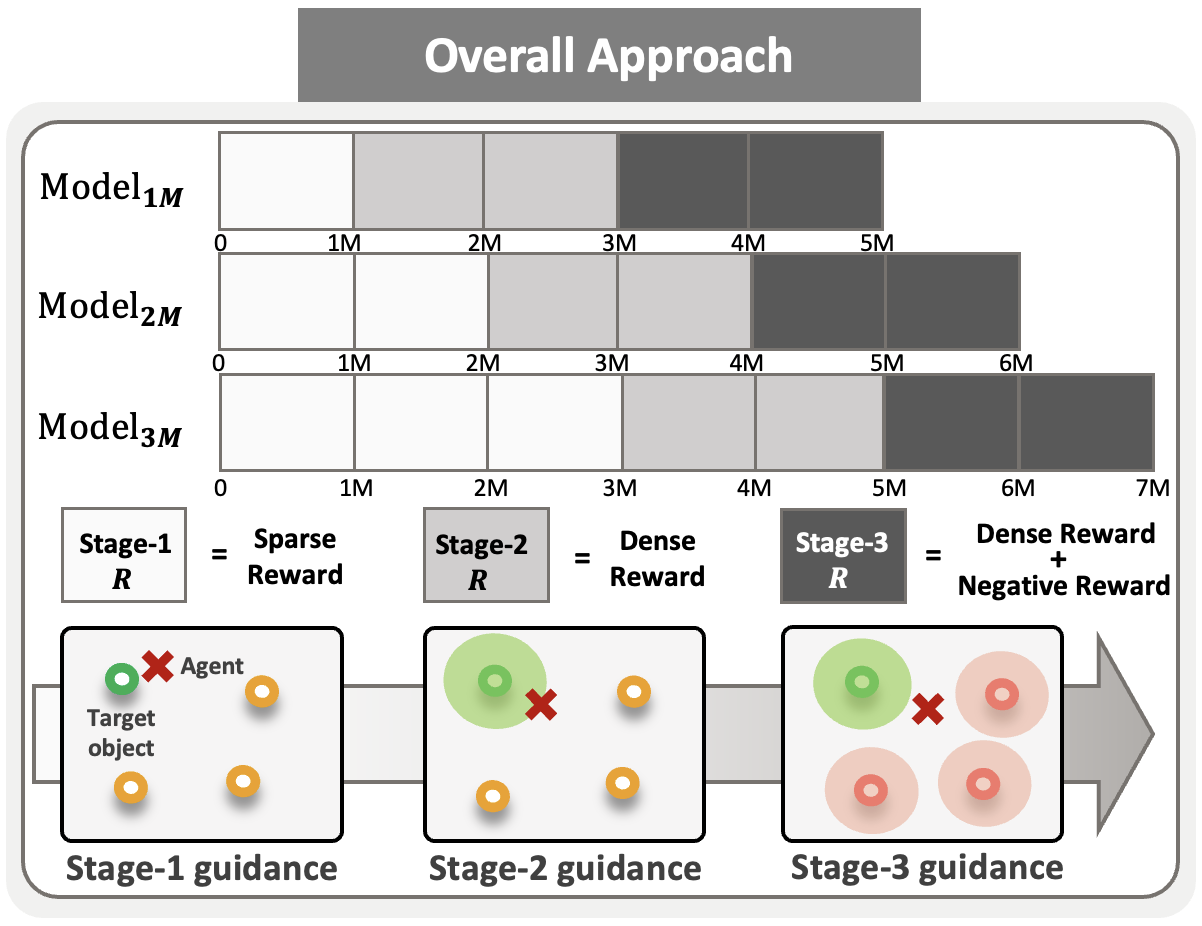}
    \vskip -0.1in
    \caption{Overview of the three multi-stage learning approach.
    In \textbf{Stage-1} guidance, a reward of +10.0 is given when the agent has only reached the object directly;
    In \textbf{Stage-2}, reward +5.0 is additionally provided when the agent is within 200 units distance from the goal;
    In \textbf{Stage-3}, in addition to the reward in Stage-2, reward -5.0 is provided when the agent is within 200 units distance from a non-target object.}
    \label{figure:vizEnvironment}
\end{figure}

Unfortunately, prior works lack a deeper investigation of what the \textit{appropriate} stimulus is for the critical period. They focused on the time duration of early experiences, and na\"ively applied reward or penalty scaling as guidance. To address this, we propose the multi-stage human guidance approach inspired by human's early cognitive developmental stages. Our key insight follows from the fact that human infants are not given task-specific goals from the start. Rather, newborn babies first explore the surroundings in an unsupervised manner, and gradually move towards supervised objectives: guidance, games, and jobs~\cite{Gibson,Piaget}. Specifically, our multi-stage guidance consists of 3 stages, as shown in Fig~\ref{figure:vizEnvironment}. In the first \textit{Free Exploration} stage, the agent is not guided. In the second stage, \textit{Guided-Play} stage, which starts from the critical period, a moderate guidance signal is given. In the final \textit{Gamification} stage, a much richer guidance is provided.

We empirically demonstrate the appropriate stimulus through multi-stage guidance richer in terms of the critical period. We first illustrate the critical period for an RL agent in ViZDoom environments~\cite{kempka2016vizdoom}. Next, we demonstrate that our multi-stage guidance notably enhances the performance compared to na\"ive reward scaling guidance at the critical period. Our study corroborates that it is crucial to design the appropriate guidance for the RL agent's critical period.

\section{Related Work}
\textbf{Critical Period in ML.} Critical period is a stage in an organism's early cognitive development during which external stimuli play a critical role in its future development~\cite{smart1991critical, Rice2000CriticalPO}. For instance, a child who grew up abroad during his or her critical period of language acquisition will have more trouble learning the native language than a child who did not~\cite{snow1978critical}. The critical period is a decisive stage for general cognitive development of language acquisition, vision and auditory learning, social relationships, and others~\cite{nickersonCP}. So a natural question is whether such phenomenon also occurs for machine learning algorithms. While a few studies discovered analogical observations in machine learning, e.g. for vision~\cite{achille2017critical} and RL~\cite{park2021toddler, Kleijn2022critical} tasks, these prior works did not investigate how to leverage such critical period for better learning performance and training speed. We introduce a multi-stage training strategy to utilize the critical period and demonstrate the most \textit{appropriate} stimulus around the critical period among sequential hand-crafted N-stage guidance.

\textbf{Curriculum Learning.} Curriculum learning (CL) is a strategy to train an ML model using tasks gradually increasing in difficulty, similar to how humans learn throughout their lifetime~\cite{Wang2021ASO}. 
Human education is organized as a curriculum; humans start with easy examples but gradually advance to more complex concepts. A natural idea is to apply such curricula to ML, not only for the resulting final performance but also for training speed~\cite{Hacohen2019OnTP} and safety~\cite{Turchetta2020SafeRL}. 
Theoretically, CL acts like unsupervised pretraining~\cite{Bengio2009CurriculumL}. It facilitates generalization
and boost the convergence~\cite{Weinshall2018CurriculumLB}. 

Starting from its original concept of easy-to-hard progression, CL generalizes to a sequence of arbitrary training criteria~\cite{Wang2021ASO}. In this view, 
our work reverses the curricula from hard to easy in terms of the sparsity of the reward, and thus can be understood as \textit{anti}-curriculum learning. Our strategy depicts the early cognitive development of a human being; unsupervised exploration dominates in the early stage, proceeding then to task-specific supervised learning~\cite{zosh2017learning}. 

Recent approaches in curriculum RL design a set of intermediate tasks (MDPs) and sequence them to increase the performance or efficiency of the RL agent in the final MDP. 
However, no prior work investigated the role of the critical period for the RL agent within the curriculum design, nor did they analyze the effects of the critical period on the curriculum. 
In contrast, our work is the first to formalize and exploit the critical period into the domain of CL. 

\section{Multi-stage RL and the Critical Period}
In this section, we formalize the critical period and the multi-stage RL within the framework of the curriculum learning. 
\subsection{Preliminaries}
RL is a field of machine learning where the agent learns through trial and error, similar to how human actually acquires skills.
It is applied to various tasks that involve sequential decision making. 
Markov Decision Process (MDP) is defined as $\langle \mathcal{S}, \mathcal{A} , \mathcal{P}, \mathcal{R}, \gamma\rangle$, where
$\mathcal{S}$ is a set of environment states, $\mathcal{A}$ is a set of possible actions, $\mathcal{P}: \mathcal{S} \times \mathcal{A} \rightarrow \Delta (\mathcal{S})$ is a transition model, $\mathcal{R}: \mathcal{S}\times \mathcal{A} \rightarrow \mathbb{R}$ is a reward function, and $\gamma$ is a discount factor. The agent in the current state $s \in \mathcal{S}$ performs the action $a \in \mathcal{A}$ according to the policy  $\pi(a|s)$, and receives the next state $s'$ and reward $\mathcal{R}(s,a)$ through the transition function $\mathcal{P}(s'|s,a)$. RL aims to obtain an optimal policy $\pi^*\in \Pi^*$ that maximizes the expected cumulative rewards $\mathbb{E}_{s_t\sim\mathcal{P},a_t\sim\pi}\left[\sum_{t=0}^{\infty} \gamma^{t}r\left(s_{t}, a_{t}\right)\right]$ with $\gamma$ applied, where $\Pi^*$ is the set of the optimal policies. 

\subsection{Curriculum Learning for RL}
It is well-known that the current RL algorithms~\cite{haarnoja2018soft,schulman2017proximal, mnih2013playing} struggle in the sparse-reward environments~\cite{aubret2019survey}.
Our approach to resolve this issue is to provide a sequential guidance as a feedback, where the guidance represents the dense reward shaping. 
To begin with, we first formulate the sparseness of the reward function $\mathcal{R}$. 
Formally, the support set of $\mathcal{R}$ is defined as follows: 
\begin{equation}
\text{supp}(\mathcal{R})=\{s\in \mathcal{S} \mid \exists a\in \mathcal{A} \,\,\, s.t. \,\,\, \mathcal{R}(s, a) \neq 0 \}.
\end{equation} 
In this work, we consider a sparse-reward environment where only a small portion of the state space is included in $\text{supp}(\mathcal{R})$, e.g., $|\text{supp}(\mathcal{R})| \ll |\mathcal{S}|$. 
We now define an \textit{anti-curriculum}, a sequence of MDPs where the reward function becomes progressively denser. 

\begin{definition}[\textit{Anti-curriculum}]
Let $\{\mathcal{M}_i\}_{i=1}^N$ be a set of MDPs where $\mathcal{M}_i = \langle \mathcal{S}, \mathcal{A}, \mathcal{P}, \mathcal{R}_i, \gamma \rangle$ for each $i$. 
We denote the sequence of MDPs $(\mathcal{M}_1, \mathcal{M}_2, \cdots, \mathcal{M}_N)$ as \textit{anti-curriculum} if $\text{supp}(\mathcal{R}_1)\subseteq \text{supp}(\mathcal{R}_2) \subseteq \cdots \subseteq \text{supp}(\mathcal{R}_N)$ and $\Pi^*_0\supseteq \Pi^*_1 \supseteq \cdots \supseteq \Pi^*_N$ hold.
Here, we denote each $\mathcal{R}_i$ as a \textit{guidance}.
\end{definition}

In the above definition, the first condition 
\begin{equation}
\text{supp}(\mathcal{R}_1)\subseteq \text{supp}(\mathcal{R}_2) \subseteq \cdots \subseteq \text{supp}(\mathcal{R}_N) 
\end{equation}
denotes that the reward function becomes more dense, i.e., the guidance becomes more explicit.
The second condition,
\begin{equation}
\Pi^*_1\supseteq \Pi^*_2 \supseteq \cdots \supseteq \Pi^*_N,
\end{equation}
constrains the optimality to be preserved during the transition of the MDPs, i.e., the optimal policies of $\mathcal{M}_i$ are also optimal in $\mathcal{M}_{i+1}$.
At a high level, the sequence of MDPs in the above definition is \textit{anti-curriculum} in the sense that the reward functions are arranged in the order of the sparsity, i.e., \textit{``sparse-to-dense"}, in contrast to the conventional curriculum learning which boost the training by arranging the tasks as \textit{``easy-to-hard"}.

The sequential guidance $(\gR_1, \gR_2, \cdots, \gR_N)$, which we call \textbf{the $N$-stage guidance}, can be designed with the inductive biases such as the domain-specific prior knowledge. 
Importantly, we hypothesize that the specific timings of the transition of the guidances highly influence the agent's performance under the given sequential guidance. 
We now introduce \textit{multi-stage RL}, the anti-curriculum learning framework with sequential guidance and \textit{the stage transition}.

\begin{definition}[Multi-stage RL]
Let $(\gR_1, \gR_2, \cdots, \gR_N)$ be a sequential guidance such that $(\gM_1, \gM_2, \cdots, \gM_N)$ is the anti-curriculum where $\mathcal{M}_i = \langle \mathcal{S}, \mathcal{A}, \mathcal{P}, \mathcal{R}_i, \gamma \rangle$ for each $i$ with the initial sparse-reward MDP $\mathcal{M}_1$.
Multi-stage RL aims to obtain an optimal policy under the MDP $\widetilde{\gM}=\langle \mathcal{S}, \mathcal{A}, \mathcal{P}, \widetilde{\gR}, \gamma \rangle$ where 
\begin{equation}
\widetilde{\gR}(s_t, a_t)=\gR_i(s_t, a_t) \quad \text{if} \quad t\in[t_{i-1}, t_{i})
\end{equation}
for each $i$, where $t_0=0$.
We denote $(t_1, t_2, \cdots, t_N)$ as the \textit{stage transitions}.
\end{definition}

\begin{figure}[t!]
    \centering
    \includegraphics[width=0.48\textwidth]{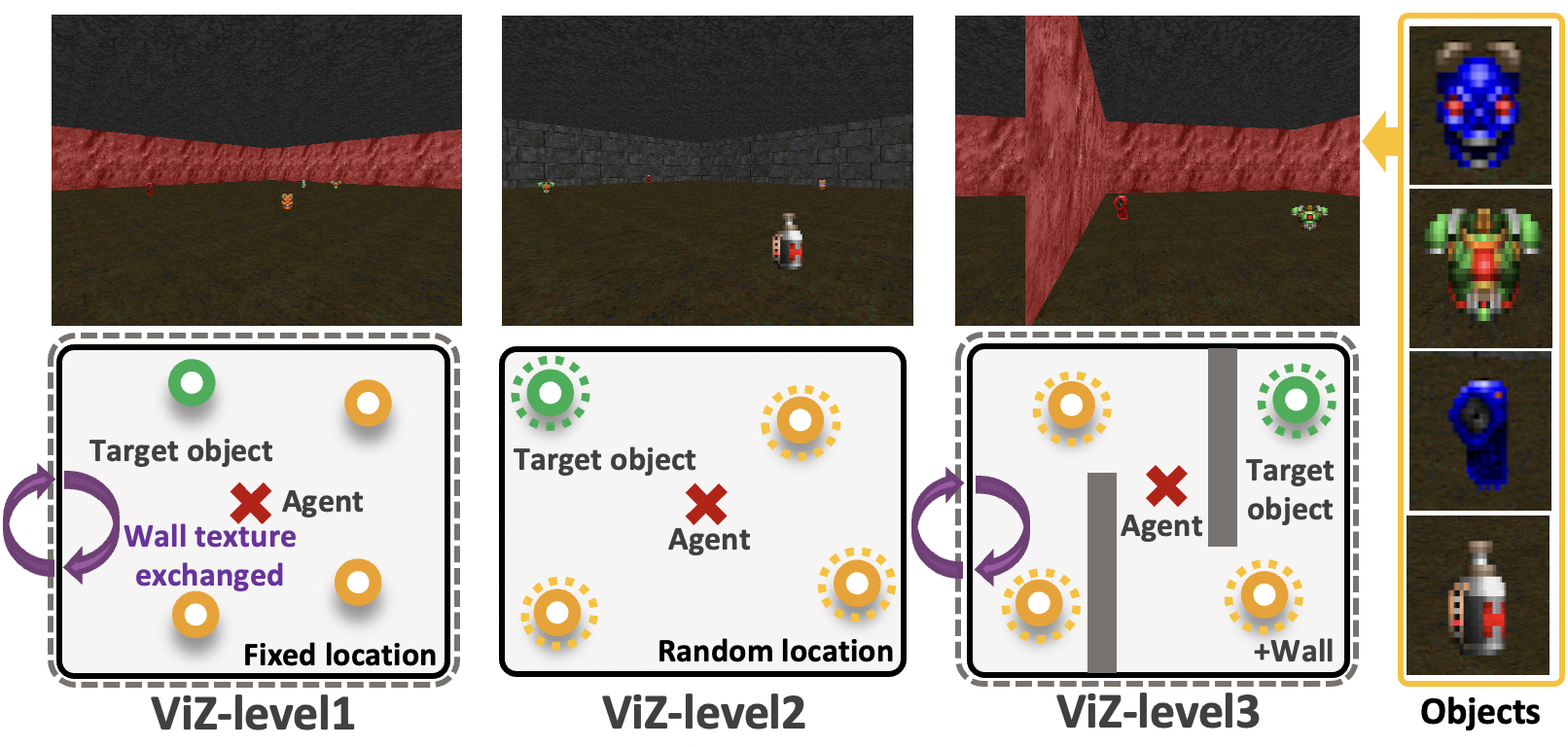}
    \vskip -0.1in
    \caption{Overview of three ViZ-level environments and objects: \textbf{level1}, fixed object locations \& changing wall texture; \textbf{level2}, random locations; \textbf{level3}, random locations \& changing wall texture \& extra walls added.}
    \label{figure:vizEnvironment2}
\end{figure}

\subsection{\textit{Anti}-Curriculum Learning and Critical Period}
In challenging sparse-reward environments, even a well-designed reward shaping guidance may be insufficient to guarantee the convergence to optimal policy.
As described earlier, the timing of the stage transition is crucial for the performance of the agent.
We hypothesize that there exists a certain period, i.e., the \textit{critical period}, which greatly affects the training when the sequential guidance is applied.
We now formalize the critical period within the framework of multi-stage RL and the sequential guidance.

\begin{definition}[Critical Period]
Let $\pi_\phi(a|s)$ be the policy parametrized with $\phi$.
Suppose that $\phi^{i}$ is updated according to an arbitrary RL algorithm $\mathcal{G}$ for each time step $i$, under the multi-stage RL framework $\widetilde{\gM}=\langle \mathcal{S}, \mathcal{A}, \mathcal{P}, \widetilde{\gR}, \gamma \rangle$ with the $N$-stage guidance $(\gR_1, \gR_2, \cdots, \gR_N)$ and the stage transition $\vt = (t_1, t_2, \cdots, t_N)$. Here, the $\epsilon$-convergence step is defined as:
\begin{equation}
L(\widetilde{\mathcal{R}}, \vt;\mathcal{G},\epsilon) = \min \{i\mid \forall s, \lvert V^{\pi_{\phi^{i}}}(s) - V^*(s)\rvert < \epsilon \}.
\end{equation}
The critical period $\vt^*$ is defined as follows:
\begin{equation}
\vt^* = \text{argmin}_{\vt}\,\, L(\widetilde{\mathcal{R}}, \vt;\mathcal{G},\epsilon).
\end{equation}
\end{definition}

Intuitively, critical period is the stage transition which obtains the optimal policies with the fastest convergence of the based RL algorithm. 
In \Cref{sec:experiments}, we empirically demonstrate critical period which shows distinguished performance among the stage transitions.

\section{Experiments}
\label{sec:experiments}
\subsection{ViZDoom 3D-Navigation Environments}
We developed 3D navigation environments using ViZDoom~\cite{kempka2016vizdoom}.
Four objects are generated in the map, one of which is the randomly specified goal object that the agent must navigate to.
Each object may spawn as one of two variants of colors or appearances.
The map size is 700 $\times$ 700 (unit distance), and the agent starts at the center of the map every episode. 
As shown in Fig~\ref{figure:vizEnvironment2}, there are three ViZ-level environments with increasing complexity.
The objects are prevented from being generated too closely to the agent, so that we can maintain a certain level of difficulty.
Arrival at a goal and non-goal object respectively grants a reward of 10.0 or -1.0 to the agent and terminates the episode.
Failing to reach any object within episode time limit (25 time steps for level 1 and 2, or 37 time steps for level 3) results in a reward of -0.1.
A reward of -0.01 is given every time step to induce active exploration.

\subsection{Results on ViZDoom Environments}

We demonstrate the impact of the multi-stage learning ranging from Free Exploration (stage-1 guidance) to Gamification (stage-3 guidance), and examine the critical periods.
For this, we measure the agent's performance across three different stage transitions: $(t_1, t_1+2M, t_1+4M)$ such that $t_1 \in \{1M, 2M, 3M\}$.
We denote these three curricula as $\widetilde{\gM}_1$, $\widetilde{\gM}_2$, and $\widetilde{\gM}_3$ respectively.
Through these experiments, we also show how an \textit{appropriate} stimulus affects the learning performance according to the critical period.
Also, we compare multi and uni-stage learning (denoted as $\widetilde{\gM}_4$), in which the richest guidance (stage-3) is given during the whole run.
A3C~\cite{mnih2016asynchronous} was used as the RL algorithm, and the means and standard deviations were recorded across three trials.

\begin{figure}[t!]
    \includegraphics[width=0.48\textwidth]{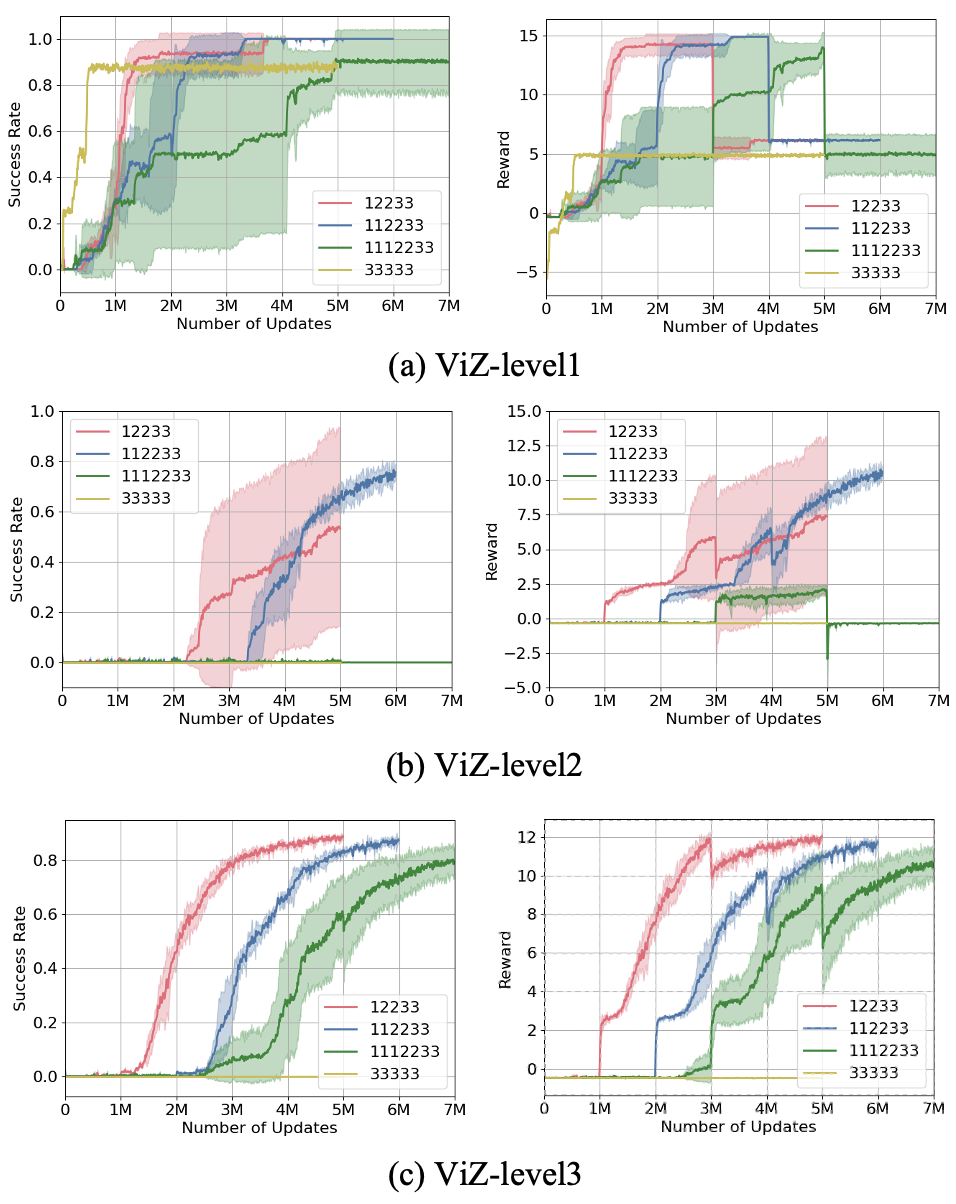}
    \vskip -0.1in
    \caption{Comparison of each $(\widetilde{\gM}_1, \widetilde{\gM}_2, \widetilde{\gM}_3, \widetilde{\gM}_4$) according to three environment levels.}
   \label{figure:vizResult}
\end{figure}
\textbf{Results on \textit{ViZ-level1} Environments.} 
In Fig~\ref{figure:vizResult}-(a), the agent reaches a perfect success rate (100\%) 
in the order of $(\widetilde{\gM}_1)$ and $(\widetilde{\gM}_2)$. 
$(\widetilde{\gM}_3)$ shows the lowest success rate (92\%). 
The uni-stage model $(\widetilde{\gM}_4)$ cannot even reach the lowest success rate of the multi-stage models (90.7\%).

\textbf{Results on \textit{ViZ-level2} Environments.}
In the results of Fig~\ref{figure:vizResult}-(b),  $(\widetilde{\gM}_2)$ shows a superior success rate (78\%). 
The $(\widetilde{\gM}_1)$ also shows a moderate performance (57\%). 
In contrast, in $(\widetilde{\gM}_3)$ and $(\widetilde{\gM}_4)$, the agent cannot solve the task properly at all (0\%).

\textbf{Results on \textit{ViZ-level3} Environments.}
For the most complex environment, all models show a larger improvement 
according to stage-2,3 guidance.
As shown in Fig~\ref{figure:vizResult}-(c), $(\widetilde{\gM}_1)$ (90\%), $(\widetilde{\gM}_2)$ (83\%) and $(\widetilde{\gM}_3)$ (78\%) exhibit best to worst performances in order.

\textbf{Overall Analysis}. 
We observe vast performance improvements during stage-2 guidance (and during stage-3 guidance to a smaller extent) in all environments, particularly with the best performing models of ViZ-level2 $(\widetilde{\gM}_2)$ and level3 $(\widetilde{\gM}_1)$.
As a result, the \textit{appropriate} stimulus, which leads to the steepest performance improvement for all ViZ-levels environments, is the stage-2 guidance. Thus, we can speculate that the periods where the best performing models are given stage-2 guidance are the critical periods
(\textit{level1}-$(\widetilde{\gM}_1)$: 1$\sim$3M/\textit{level2-$(\widetilde{\gM}_2)$}: 2$\sim$4M/ \textit{level3-$(\widetilde{\gM}_1)$}: 1$\sim$3M). 

We also observe that standard deviations are high when stage-2 guidance is given outside the critical period. Therefore, we believe that a proper amount of free exploration stage (stage-1) is essential to stable learning so that the agent fully learns an optimal trajectory. For uni-stage $(\mathcal{M}_4)$, even if it is the richest reward, the agent fails completely in level 2 and 3. However, we found that stage-3 guidance transforms into a beneficial guidance through stage transition $(t_2)$ in multi-stage approach.    

\section{Conclusion}
Appropriate reward shaping is a crucial performance factor for successful learning in MDP. However, it is imperfect due to human cognitive bias. Therefore, to find a beneficial reward shaping more clearly, we introduce multi-stage RL and demonstrate the appropriate stimulus.
Our work is the first to formalize and exploit critical period within the framework of curriculum RL. 
Our empirical experiments demonstrate the effectiveness of multi-stage learning near the critical period in terms of the performance, efficiency, and stability of the training for the RL agent. While feedback types within anti-curriculum learning have been rather underexamined, our multi-stage RL framework leads to numerous relevant open problems. For instance, the generalization performance of multi-stage RL to unseen tasks is worth examining. Also, the effects of various combinations of guidance types need investigation.

\section*{Acknowledgements}

This work was partly supported by the Institute of Information \& Communications Technology Planning \& Evaluation (2015-0-00310-SW.StarLab/10\%, 2019-0-01371-BabyMind/20\%, 2021-0-02068-AIHub/10\%, 2021-0-01343-GSAI/10\%, 2022-0-00951-LBA/10\%, 2022-0-00166-PICA/20\%) and the NRF of Korea (2021R1A2C10-10970/20\%) grant funded by the Korean government.


\nocite{langley00}

\bibliography{bibliography}

\begin{thebibliography}{23}
\providecommand{\natexlab}[1]{#1}
\providecommand{\url}[1]{\texttt{#1}}
\expandafter\ifx\csname urlstyle\endcsname\relax
  \providecommand{\doi}[1]{doi: #1}\else
  \providecommand{\doi}{doi: \begingroup \urlstyle{rm}\Url}\fi

\bibitem[Achille et~al.(2017)Achille, Rovere, and Soatto]{achille2017critical}
Achille, A., Rovere, M., and Soatto, S.
\newblock Critical learning periods in deep neural networks.
\newblock \emph{arXiv preprint arXiv:1711.08856}, 2017.

\bibitem[Aubret et~al.(2019)Aubret, Matignon, and Hassas]{aubret2019survey}
Aubret, A., Matignon, L., and Hassas, S.
\newblock A survey on intrinsic motivation in reinforcement learning.
\newblock \emph{arXiv preprint arXiv:1908.06976}, 2019.

\bibitem[Bengio et~al.(2009)Bengio, Louradour, Collobert, and
  Weston]{Bengio2009CurriculumL}
Bengio, Y., Louradour, J., Collobert, R., and Weston, J.
\newblock Curriculum learning.
\newblock In \emph{ICML '09}, 2009.

\bibitem[Gibson(1988)]{Gibson}
Gibson, E.~J.
\newblock Exploratory behavior in the development of perceiving, acting, and
  the acquiring of knowledge.
\newblock \emph{Annual review of psychology}, 39\penalty0 (1):\penalty0 1--42,
  1988.

\bibitem[Haarnoja et~al.(2018)Haarnoja, Zhou, Abbeel, and
  Levine]{haarnoja2018soft}
Haarnoja, T., Zhou, A., Abbeel, P., and Levine, S.
\newblock Soft actor-critic: Off-policy maximum entropy deep reinforcement
  learning with a stochastic actor.
\newblock In \emph{International conference on machine learning}, pp.\
  1861--1870. PMLR, 2018.

\bibitem[Hacohen \& Weinshall(2019)Hacohen and Weinshall]{Hacohen2019OnTP}
Hacohen, G. and Weinshall, D.
\newblock On the power of curriculum learning in training deep networks.
\newblock \emph{ArXiv}, abs/1904.03626, 2019.

\bibitem[Kempka et~al.(2016)Kempka, Wydmuch, Runc, Toczek, and
  Ja{\'s}kowski]{kempka2016vizdoom}
Kempka, M., Wydmuch, M., Runc, G., Toczek, J., and Ja{\'s}kowski, W.
\newblock Vizdoom: A doom-based ai research platform for visual reinforcement
  learning.
\newblock In \emph{2016 IEEE conference on computational intelligence and games
  (CIG)}, pp.\  1--8. IEEE, 2016.

\bibitem[Kleijn et~al.(2022)Kleijn, Sen, and Kachergis]{Kleijn2022critical}
Kleijn, de~Kleijn, R., Sen, D., and Kachergis, G.
\newblock A critical period for robust curriculum-based deep reinforcement
  learning of sequential action in a robot arm.
\newblock \emph{Topics in Cognitive Science}, 2022.

\bibitem[Mnih et~al.(2013)Mnih, Kavukcuoglu, Silver, Graves, Antonoglou,
  Wierstra, and Riedmiller]{mnih2013playing}
Mnih, V., Kavukcuoglu, K., Silver, D., Graves, A., Antonoglou, I., Wierstra,
  D., and Riedmiller, M.
\newblock Playing atari with deep reinforcement learning.
\newblock \emph{arXiv preprint arXiv:1312.5602}, 2013.

\bibitem[Mnih et~al.(2016)Mnih, Badia, Mirza, Graves, Lillicrap, Harley,
  Silver, and Kavukcuoglu]{mnih2016asynchronous}
Mnih, V., Badia, A.~P., Mirza, M., Graves, A., Lillicrap, T., Harley, T.,
  Silver, D., and Kavukcuoglu, K.
\newblock Asynchronous methods for deep reinforcement learning.
\newblock In \emph{International conference on machine learning}, pp.\
  1928--1937. PMLR, 2016.

\bibitem[Nickerson(2021)]{nickersonCP}
Nickerson, C.
\newblock Critical period in brain development and childhood learning.
\newblock In \emph{Simply Psychology}, 2021.

\bibitem[Park et~al.(2021)Park, Park, Oh, Lee, Lee, Lee, and
  Zhang]{park2021toddler}
Park, J., Park, K., Oh, H., Lee, G., Lee, M., Lee, Y., and Zhang, B.-T.
\newblock Toddler-guidance learning: Impacts of critical period on multimodal
  ai agents.
\newblock In \emph{Proceedings of the 2021 International Conference on
  Multimodal Interaction}, pp.\  212--220, 2021.

\bibitem[Piaget \& Cook(1952)Piaget and Cook]{Piaget}
Piaget, J. and Cook, M.
\newblock \emph{The origins of intelligence in children}, volume~8.
\newblock International Universities Press New York, 1952.

\bibitem[Rice \& Barone(2000)Rice and Barone]{Rice2000CriticalPO}
Rice, D.~C. and Barone, S.
\newblock Critical periods of vulnerability for the developing nervous system:
  evidence from humans and animal models.
\newblock \emph{Environmental Health Perspectives}, 108:\penalty0 511 -- 533,
  2000.

\bibitem[Schulman et~al.(2017)Schulman, Wolski, Dhariwal, Radford, and
  Klimov]{schulman2017proximal}
Schulman, J., Wolski, F., Dhariwal, P., Radford, A., and Klimov, O.
\newblock Proximal policy optimization algorithms.
\newblock \emph{arXiv preprint arXiv:1707.06347}, 2017.

\bibitem[Smart(1991)]{smart1991critical}
Smart, J.~L.
\newblock Critical periods in brain development.
\newblock In \emph{The childhood environment and adult disease: Ciba Foundation
  Symposium}, volume 156, pp.\  109--28, 1991.

\bibitem[Snow \& Hoefnagel-H{\"o}hle(1978)Snow and
  Hoefnagel-H{\"o}hle]{snow1978critical}
Snow, C.~E. and Hoefnagel-H{\"o}hle, M.
\newblock The critical period for language acquisition: Evidence from second
  language learning.
\newblock \emph{Child development}, pp.\  1114--1128, 1978.

\bibitem[Takesian \& Hensch(2013)Takesian and Hensch]{Takesian2013BalancingPA}
Takesian, A.~E. and Hensch, T.~K.
\newblock Balancing plasticity/stability across brain development.
\newblock \emph{Progress in brain research}, 207:\penalty0 3--34, 2013.

\bibitem[Turchetta et~al.(2020)Turchetta, Kolobov, Shah, Krause, and
  Agarwal]{Turchetta2020SafeRL}
Turchetta, M., Kolobov, A., Shah, S., Krause, A., and Agarwal, A.
\newblock Safe reinforcement learning via curriculum induction.
\newblock \emph{ArXiv}, abs/2006.12136, 2020.

\bibitem[Wang et~al.(2021)Wang, Chen, and Zhu]{Wang2021ASO}
Wang, X., Chen, Y., and Zhu, W.
\newblock A survey on curriculum learning.
\newblock \emph{IEEE transactions on pattern analysis and machine
  intelligence}, PP, 2021.

\bibitem[Weinshall \& Cohen(2018)Weinshall and
  Cohen]{Weinshall2018CurriculumLB}
Weinshall, D. and Cohen, G.
\newblock Curriculum learning by transfer learning: Theory and experiments with
  deep networks.
\newblock \emph{ArXiv}, abs/1802.03796, 2018.

\bibitem[White et~al.(2013)White, Hutka, Williams, and
  Moreno]{white2013learning}
White, E.~J., Hutka, S.~A., Williams, L.~J., and Moreno, S.
\newblock Learning, neural plasticity and sensitive periods: implications for
  language acquisition, music training and transfer across the lifespan.
\newblock \emph{Frontiers in systems neuroscience}, 7:\penalty0 90, 2013.

\bibitem[Zosh et~al.(2017)Zosh, Hopkins, Jensen, Liu, Neale, Hirsh-Pasek,
  Solis, and Whitebread]{zosh2017learning}
Zosh, J.~N., Hopkins, E.~J., Jensen, H., Liu, C., Neale, D., Hirsh-Pasek, K.,
  Solis, S.~L., and Whitebread, D.
\newblock \emph{Learning through play: a review of the evidence}.
\newblock LEGO Fonden Billund, Denmark, 2017.

\end{thebibliography}
\bibliographystyle{icml2022}



\end{document}